\documentclass[10pt,journal,compsoc]{IEEEtran}

\ifCLASSOPTIONcompsoc
  \usepackage[nocompress]{cite}
\else
  \usepackage{cite}
\fi
\ifCLASSINFOpdf
\else
\fi
\usepackage{amsmath}
\usepackage{amssymb}
\usepackage{mathtools}
\usepackage{amsthm}

\usepackage{amsfonts}
\usepackage{multirow}
\usepackage{multicol}

\usepackage{booktabs}
\usepackage{color}
\usepackage{algorithm}
\usepackage{algorithmic}

\usepackage{graphicx}
\usepackage{subfig}

\newtheorem{theorem}{Theorem}
\newtheorem{lemma}{Lemma}

\newtheorem{assumption}{Assumption}
\newtheorem{remark}{Remark}

\begin{document}
%
\title{ Compositional federated learning: Applications in distributionally robust averaging and meta learning }
%
%
%
%

\author{Feihu~Huang,~~Junyi~Li 
\IEEEcompsocitemizethanks{\IEEEcompsocthanksitem Feihu Huang is with College of Computer Science and Technology,
Nanjing University of Aeronautics and Astronautics, Nanjing, China.
E-mail: huangfeihu2018@gmail.com
\IEEEcompsocthanksitem Junyi~Li is with Department of Electrical and Computer Engineering, University of Pittsburgh, USA.
E-mail: junyili.ai@gmail.com

}}

%
%

\markboth{Journal of \LaTeX\ Class Files,~Vol.~14, No.~8, August~2015}%
{Shell \MakeLowercase{\textit{et al.}}: Bare Advanced Demo of IEEEtran.cls for IEEE Computer Society Journals}
%



\IEEEtitleabstractindextext{%
\begin{abstract}
In the paper, we propose an effective and efficient Compositional Federated Learning (ComFedL) algorithm
for solving a new compositional Federated Learning (FL) framework,
which frequently appears in many data mining and machine learning problems with a hierarchical structure such as
distributionally robust FL
and model-agnostic meta learning (MAML).
Moreover, we study the convergence analysis of our ComFedL algorithm under some mild conditions,
and prove that it achieves a convergence rate of $O(\frac{1}{\sqrt{T}})$,
where $T$ denotes the number of iteration.
To the best of our knowledge, our new Compositional FL framework is the
first work to bridge federated learning with composition stochastic optimization.
In particular, we first transform the distributionally robust FL (i.e., a minimax optimization problem) into
a simple composition optimization problem by using KL divergence regularization.
At the same time, we also first transform the distribution-agnostic MAML problem (i.e., a minimax optimization problem)
into a simple yet effective composition optimization problem.
Finally, we apply two popular machine learning tasks,
i.e., distributionally robust FL and MAML to demonstrate
the effectiveness of our algorithm.
\end{abstract}

}

\maketitle

\IEEEdisplaynontitleabstractindextext

%
\IEEEpeerreviewmaketitle

\ifCLASSOPTIONcompsoc
\IEEEraisesectionheading{\section{Introduction}\label{sec:introduction}}
\else
\section{Introduction}
\label{sec:introduction}
\fi

%
%
%
%

\IEEEPARstart{I}{n} this paper, we study a new compositional Federated Learning (FL) framework that is equivalent to solve the following composition optimization
problem, defined as
\begin{align}  \label{eq:1}
 \min_{w \in \mathbb{R}^d}\frac{1}{n}\sum_{i=1}^n \mathbb{E}_{\zeta\sim \tilde{\mathcal{D}}_i}\big[ g^i\big(\mathbb{E}_{\xi\sim \mathcal{D}_i}[f^i(w;\xi)];\zeta\big)\big],
\end{align}
where $n$ denotes the number of nodes (devices), and $y^i=f^i(w)=\mathbb{E}_{\xi\sim \mathcal{D}_i}[f^i(w;\xi)]: \mathbb{R}^d\rightarrow \mathbb{R}^p$
denotes the inner function (or mapping) and $g^i(y^i)=\mathbb{E}_{\zeta\sim \tilde{\mathcal{D}}_i}[g^i(y^i;\zeta)]: \mathbb{R}^p \rightarrow \mathbb{R}$
denotes the outer function. Here let $F(w) := \frac{1}{n}\sum_{i=1}^n F^i(w)$, and
$F^i(w)=g^i(f^i(w)): \mathbb{R}^d\rightarrow \mathbb{R}$ is smooth but possibly nonconvex function.
$\{\mathcal{D}_i,\tilde{\mathcal{D}}_i\}$ denote two data distributions on $i$-th device, and $\{\mathcal{D}_i\}_{i=1}^n$ are \textbf{not identical},
similar for $\{\tilde{\mathcal{D}}_i\}_{i=1}^n$.
In our new FL framework, clients collaboratively learn a model,
but the raw data in each device is never shared with the server and other devices as in the existing FL framework~\cite{mcmahan2017communication,kairouz2019advances}.
Thus, it is also helpful to protect data privacy as in the existing FL framework.

A key difference between our new FL framework and
the existing FL framework is that each device includes two different data distributions $\{\mathcal{D}_i,\tilde{\mathcal{D}}_i\}$
in our new FL framework, while each device only includes one data distribution $\mathcal{D}_i$
in existing FL framework. Another key difference is that our loss function is two-level composition function,
while the loss function in the existing FL framework is only a single-level function.
Thus, our new FL framework can effectively be applied to many machine learning problems with a hierarchical structure
such as distributionally robust FL and model-agnostic meta learning.

\subsection{ Applications }
\textbf{1). Distributionally Robust Federated Learning.}
Federated learning (FL)~\cite{mcmahan2017communication,kairouz2019advances} is a popular learning paradigm in machine learning for training a centralized model
using data distributed over a network of devices.
In general, FL solves the following distributed optimization problem:
\begin{align} \label{eq:2}
 \min_{w \in \mathbb{R}^d} \sum_{i=1}^n r_i f^i(w), \quad f^i(w):=\mathbb{E}_{\xi \sim \mathcal{D}_i}[\ell_i(w;\xi)]
\end{align}
where $r_i\in (0,1)$ denotes the proportion of $i$-th device in the entire model. Here $\ell_i(w;\xi)$ is
the loss function on $i$-th device, and $\mathcal{D}_i$ denotes the data distribution on $i$-th device.
In FL, the data distributions $\{\mathcal{D}_i\}_{i=1}^n$ generally are different. The goal of FL is to
learn a global variable $w$ based on these heterogeneous data from different data distributions.

To tackle the data heterogeneity concern in FL,
some robust FL algorithms \cite{mohri2019agnostic,reisizadeh2020robust,deng2020distributionally} have been proposed.
Specifically, the robust FL mainly focuses on the following agnostic (distributionally robust) empirical loss problem
\begin{align} \label{eq:3}
 \min_{w \in \mathbb{R}^d} \max_{r\in \Lambda_n} \sum_{i=1}^n r_i f^i(w),
\end{align}
where $\Lambda_n = \{ r\in \mathbb{R}^n_+ : \sum_{i=1}^nr_i=1, \ r_i\geq 0\}$ is
a $n$-dimensional simplex. In fact, these robust FL algorithms find a global variable $w$
from the worst-case loss, so the obtained variable $w$ is robust to the data heterogeneity.
In the paper, we further introduce a regularized agnostic empirical loss problem as follows:
\begin{align} \label{eq:4}
 \min_{w \in \mathbb{R}^d} \max_{r\in \Lambda_n} \sum_{i=1}^n r_i f^i(w) - \gamma\phi(r,1/n),
\end{align}
where $\gamma >0$ is a regularization parameter, and $\phi(r,1/n)$
is a divergence measure between $r_i$ for all $i\in [n]$ and uniform probability $1/n$.
Our motivation is that this penalty $\phi(r,1/n)$ ensures that the proportion $r_i$ is not far away from $1/n$, i.e.,
we still equally treat each local dataset and model in training the whole model.
When consider the KL divergence $\phi(r,1/n)= \sum_{i=1}^n r_i\log(nr_i)$,
by exactly maximizing over $r\in \Lambda_n$, the above minimax problem (\ref{eq:4}) is equivalent to a composition problem, defined as
\begin{align} \label{eq:5}
 \min_{w \in \mathbb{R}^d} \gamma\log\bigg( \frac{1}{n}\sum_{i=1}^n \exp\big(f^i(w)/\gamma\big)\bigg),
\end{align}
which can be obtained from Lemma \ref{lem:1} given in the following Section \ref{sec:4}.
Since the function $\log(\cdot)$ is monotonically increasing, we can solve the following problem instead of the above problem
(\ref{eq:5}), defined as
\begin{align} \label{eq:6}
 \min_{w \in \mathbb{R}^d} \frac{1}{n}\sum_{i=1}^n g\big(f^i(w)/\gamma\big),
\end{align}
where $g(\cdot)=\exp(\cdot/\gamma)$. In fact, we can apply the other monotonically increasing functions
instead of the function $g(\cdot)=\exp(\cdot/\gamma)$ in the problem \eqref{eq:6}.
Clearly, the problem (\ref{eq:6}) is a special case of the above problem (\ref{eq:1}).

In fact, our ComFedL framework \eqref{eq:6} exponentially scales the loss of each device. Specifically, $f^i(w)$ denotes loss of the $i$-th device in
the original FL problem \eqref{eq:2}. In our ComFedL problem \eqref{eq:6},
while we have a new loss function $\exp(f^i(w)/\gamma)$, and its gradient in the form of $(\exp(f^i(w)/\gamma)/\gamma) \nabla f^i(w)$.
We can view our ComFedL as re-weighting the clients based on its current loss, higher loss leads to higher weights, so that we can learn a fairer model.

\textbf{2). Model-Agnostic Meta Learning.}
Meta Learning is a powerful learning paradigm for learning the optimal model properties to improve model performances
with more experiences, i.e., learning to learn \cite{andrychowicz2016learning}. Model-Agnostic Meta Learning (MAML) \cite{finn2017model} is a popular meta-learning method, which is to learn a good initialization for a gradient-based update. The goal of MAML is to find a common initialization that can adapt to a desired model for a set of new tasks after taking several gradient descent steps.
Specifically, we consider a set of tasks collected in $\mathcal{M}:=\{1,2,\cdots,n\}$ drawn from a certain task distribution.
Then we find such
initialization by solving the following one-step MAML problem
\begin{align} \label{eq:7}
 \min_{w\in \mathbb{R}^d} \frac{1}{n}\sum_{i=1}^n f_i\big(w-\eta\nabla f_i(w)\big),
\end{align}
where $f_i(w)=\mathbb{E}_{\xi\sim \mathcal{D}_i}[f(w;\xi)]$, and random variable $\xi$ follows the unknown distribution $\mathcal{D}_i$, and $\eta >0$ is stepsize.
Let $g^i(y^i)=f_i(y^i)$ and $y^i=f^i(w)=w-\eta\nabla f_i(w)$, the above problem (\ref{eq:7}) is a special case of the
above composition problem (\ref{eq:1}).

Recently, \cite{fallah2020personalized} applied (\ref{eq:7}) to perform personalized Federated Learning, where every client is viewed as a task. More recently, \cite{collins2020distribution} pointed out some drawbacks such as poor worst-case performance and unfairness of the above MAML problem (\ref{eq:7}). To overcome these drawbacks, \cite{collins2020distribution} proposed a task-robust MAML (TR-MAML) model, defined as
\begin{align} \label{eq:8}
    \min_{w\in \mathbb{R}^d} \max_{p\in \Delta_n} \sum_{i=1}^n p_i f_i\big(w-\eta\nabla f_i(w)\big)
\end{align}
where $p_i$ denotes the probability associated with $i$-th task, and $p=(p_1,\cdots,p_n)$, and
$\Delta_n = \{p\in \mathbb{R}^n_+| \sum_{i=1}^np_i=0, p_i\geq 0\}$ is the standard simplex in $\mathbb{R}^n_+$.
Following the above distributionally robust FL, we can also add a KL divergence $\phi(p,1/n)=\sum_{i=1}^np_i\log(np_i)$
to the above problem (\ref{eq:8}).
Then we obtain a regularized TR-MAML model, defined as
\begin{align} \label{eq:9}
\min_{w\in \mathbb{R}^d} \max_{p\in \Delta_n} \sum_{i=1}^n p_i f_i\big(w-\eta\nabla f_i(w)\big) - \gamma \phi(p,1/n)
\end{align}
where tuning parameter $\gamma>0$.
Since the above formation (\ref{eq:7}) can be regarded as
a personalized federated learning problem \cite{fallah2020personalized}, the formation (\ref{eq:9})
can be also regarded as a distributionally robust personalized federated learning problem.
Similarly, following the above problem (\ref{eq:6}), by exactly
maximizing over $p\in \Delta_n$,
we can also solve the following composition problem instead of the above minimax problem (\ref{eq:9}), defined as
\begin{align} \label{eq:10}
 \min_{w \in \mathbb{R}^d} \frac{1}{n}\sum_{i=1}^n \exp\big(f_i\big(w-\eta\nabla f_i(w)\big)/\gamma\big).
\end{align}
Let $g^i(y^i)=\exp(f_i(y^i)/\gamma)$ and $y^i=f^i(w)=w-\eta\nabla f_i(w)$, the above problem (\ref{eq:10}) also is a special case of the
above problem (\ref{eq:1}).

\subsection{ Contributions }
Our main \textbf{contributions} can be summarized as follows:
\begin{itemize}
\setlength{\itemsep}{0pt}
\item[1)] We introduce a novel compositional federated learning framework that can be applied in many popular machine learning problems such as distributionally robust FL and (distribution-agnostic) MAML.
Moreover, we propose an effective and efficient Compositional Federated Learning (ComFedL) algorithm for solving this new compositional FL framework.
\item[2)] We provide a convergence analysis framework for our algorithm. Specifically,
we prove that our ComFedL algorithm reaches a convergence rate of $O(\frac{1}{\sqrt{T}})$, where $T$ denotes the number of iteration.
\item[3)] To the best of our knowledge, our compositional FL is the first work to bridge federated learning with composition stochastic optimization.
In particular, we first transform the distributionally robust FL (i.e., a minimax problem) into a simple yet effective composition problem by using KL divergence regularization (please see the above problem (\ref{eq:6})).
\item[4)] Since the above problem (\ref{eq:9})
can be regarded as a distributionally robust personalized FL problem, we are the first study the distributionally robust personalized FL based on composition optimization.
At the same time, we also first transform the distribution-agnostic MAML problem
into a simple yet effective composition problem (please see the above problem (\ref{eq:10})).
\item[5)] Extensive experimental results on distributionally robust FL and MAML demonstrate the efficiency of our algorithm.
\end{itemize}
\section{Related Works}
In this section, we review federated learning, composition optimization and
model-agnostic meta learning.

\subsection{ Federated Learning }
\begin{figure}[!t]
 \centering
 \includegraphics[width=0.32\textwidth]{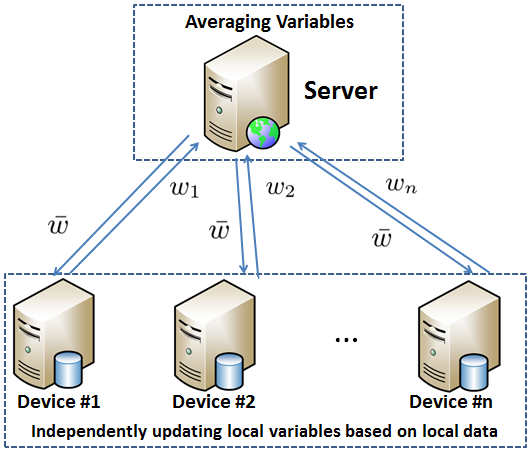}\\
 \caption{A star-network of FL system includes a server and multiple devices. In the basic
 FL problem \eqref{eq:2}, the server averages the local model variables $\bar{w}=\sum_{i=1}^n r_i w_i$, and sends $\bar{w}$ to each device;
 Each device update the local variable $w_i$ for $i\in [n]$, and then sent it to the server. }
 \label{fig:1}
 \vspace*{-8pt}
\end{figure}

Federated Learning (FL)~\cite{mcmahan2017communication,kairouz2019advances} has become a useful paradigm
in large-scale machine learning applications such as automatic disease diagnosis \cite{yang2021flop}, where the data remains distributed over a large number of clients such as network sensors or mobile phones.
The key point of FL is that multiple clients such as edge devices cooperate to learn a global model
and raw client data is never shared with the server and the other clients ( Please see Figure \ref{fig:1}). Thus, it is very helpful to protect data privacy.
FedAvg \cite{mcmahan2017communication} is the first FL algorithm, which builds on the local stochastic gradient descent (SGD) \cite{haddadpour2019trading,haddadpour2019local, woodworth2020local, khaled2020tighter,liang2019variance, yu2019parallel, acar2020federated}.
Specifically, it performs multiple SGDs in the available clients before communicating to the server,
which can reduce the total amount of communication required, but can lead to client drift \cite{karimireddy2019scaffold, li2019convergence, sahu2018convergence, zhao2018federated}.
To deal with these issues of client drift and stability, some variants of FedAvg have been proposed.
For example, \cite{li2018federated} applied a regularization term in the client objectives towards the broadcast model to reduce the client drift.
\cite{hsu2019measuring,karimireddy2019scaffold,wang2019slowmo} used the momentum technique on the server to control variates. \cite{karimireddy2020mime} applied the momentum techniques on both the clients and server to control variates.
In addition, to improve flexibility
and scalability of FL, \cite{xie2019asynchronous} studied a new asynchronous federated learning based on asynchronous training.

Recently, some robust FL methods \cite{mohri2019agnostic,reisizadeh2020robust,deng2020distributionally} have been proposed to deal with the
data heterogeneity concern in FL. Specifically, these robust FL methods mainly learning the worst-case loss by solving a minimax optimization problems. To incorporate personalization in FL, more recently
some personalized federated learning models
\cite{chen2021bridging, li2021ditto, jiang2019improving,fallah2020personalized,deng2020adaptive, dinh2020personalized, chen2018federated} have been developed and studied.
For example, Ditto \cite{li2021ditto} is a recently proposed personalized FL algorithm where a regularized local model is learnt for each client.

\subsection{ Composition Stochastic Optimization }
Composition stochastic optimization has been widely applied to
many applications such as reinforcement learning \cite{wang2017accelerating},
model-agnostic meta Learning \cite{tutunov2020compositional} and risk management \cite{wang2017accelerating}.
Some compositional gradient methods have recently been proposed to solve these composition optimization problems.
For example, stochastic compositional gradient methods \cite{wang2017stochastic,wang2017accelerating,ghadimi2020single} have been proposed
to solve these problems.
Subsequently, some variance-reduced compositional algorithms \cite{huo2018accelerated,lin2018improved,zhang2019multi}
have been proposed for composition optimization.
More recently, \cite{tutunov2020compositional,chen2020solving} presented a class of adaptive compositional
gradient methods.
\subsection{Model-Agnostic Meta Learning}
Model-Agnostic Meta Learning (MAML) is an effective learning framework that learns a good initialization from prior experiences to fast adaptation in
new tasks \cite{finn2017model,finn2019online}.
MAML has been widely used to various applications such as deep learning \cite{finn2017model}, reinforcement learning
\cite{liu2019taming,fallah2020provably, fallah2020convergence} and personalized federated learning \cite{fallah2020personalized}. For example, \cite{fallah2020provably} has proposed stochastic gradient meta-reinforcement learning, which can be regarded as a variant of the MAML method.
\cite{fallah2020personalized} has studied a personalized variant of federated learning from the MAML view.

Due to solving MAML requires information on the stochastic Hessian matrix, some Hessian-free methods \cite{balcan2019provable,song2019maml} have been recently proposed to reduce the cost of computing Hessian matrix. At the same time, \cite{ji2020multi} studied the general multi-step MAML instead of one-step one to improve performance. More recently, \cite{tutunov2020compositional,chen2020solving} have applied the composition optimization to the MAML by using the compositional structure of MAML.

\textbf{Notations:}
$\|\cdot\|$ denotes the $\ell_2$ norm for vectors and spectral norm for matrices.
$\langle x,y\rangle$ denotes the inner product of two vectors $x$ and $y$. $M^T$ denotes transpose of matrix $M$.
For two sequences $\{a_n\}$ and $\{b_n\}$, we denote $a_n=O(b_n)$ if $a_n\leq Cb_n$ for constant $C>0$.

\section{ Compositional Federated Learning }
In this section, we propose an effective and efficient compositional federated learning (ComFedL) algorithm
to solve the problem (\ref{eq:1}) based on compositional gradient descent iteration.
The pseudo code of ComFedL algorithm is given in Algorithm \ref{alg:1}.

\begin{algorithm}[tb]
\caption{ Compositional Federated Learning (\textbf{ComFedL}) Algorithm }
\label{alg:1}
\begin{algorithmic}[1] 
\STATE {\bfseries Input:}  Synchronization gap $\tau$, number of outer iterations $S$, learning rate $\eta$,
and initial parameter $\bar{w}_0 \in \mathbb{R}^d$; \\
\FOR{$s = 0, 1, \ldots, S-1$}
\STATE Server samples a mini-batch devices $\mathcal{C}_s \subset [n]$ with $|\mathcal{C}_s|=m$;
\STATE Server broadcasts $\bar{w}_s$ to all devices $i\in \mathcal{C}_s$;
\FOR{Devices $i\in \mathcal{C}_s$ parallel}
\STATE Device sets $w^i_{s,0} = \bar{w}_s$;
\FOR{$t = 0, 1, \cdots, \tau-1$}
\STATE Device draws a mini-batch samples $\mathcal{B}^i_t \subset \mathcal{D}_i$ with $|\mathcal{B}^i_t|=b$;
\STATE Device computes $f^i_{\mathcal{B}^i_t}(w^i_{s,t})=\frac{1}{b}\sum_{j\in \mathcal{B}^i_t}f^i(w^i_{s,t};\xi_j)$ and $\nabla f^i_{\mathcal{B}^i_t}(w^i_{s,t})=\frac{1}{b}\sum_{j\in \mathcal{B}^i_t}\nabla f^i(w^i_{s,t};\xi_j)$; \\
\STATE Device draws a mini-batch samples $\tilde{\mathcal{B}}^i_t \subset \tilde{\mathcal{D}}_i$ with $|\tilde{\mathcal{B}}^i_t|=b_1$; \\
\STATE Device computes $\nabla g^i_{\tilde{\mathcal{B}}^i_t}(f^i_{\mathcal{B}^i_t}(w^i_{s,t}))= \frac{1}{b_1}\sum_{j\in \tilde{\mathcal{B}}^i_t} \nabla g^i(f^i_{\mathcal{B}^i_t}(w^i_{s,t});\zeta_j)$; \\
\STATE $u^i_{s,t} = \nabla g^i_{\tilde{\mathcal{B}}^i_t}(f^i_{\mathcal{B}^i_t}(w^i_{s,t}))^T\nabla f^i_{\mathcal{B}^i_t}(w^i_{s,t})$;
\STATE $w^i_{s,t+1} = w^i_{s,t} - \eta u^i_{s,t}$;
\ENDFOR
\STATE Device sends $w^i_{s,\tau}$ back to server; \\
\ENDFOR
\STATE Server computes $\bar{w}_{s+1} = \frac{1}{m}\sum_{i\in \mathcal{C}_s} w^i_{s,\tau}$; \\
\ENDFOR
\STATE {\bfseries Output:} Final solution $w_T$.
\end{algorithmic}
\end{algorithm}

For solving the problem (\ref{eq:1}), we should compute the gradient of composition function $F^i(w)=g^i(f^i(w))$,
defined as
\begin{align}
 \nabla F^i (w) = \big(\nabla g^i(f^i(w))\big)^T\nabla f^i(w).
\end{align}
Since the inner function (or mapping) $y^i=f^i(w)=\mathbb{E}_{\xi\sim \mathcal{D}_i}f^i(w;\xi)$ and the outer function $g^i(y^i)=\mathbb{E}_{\zeta\sim \tilde{\mathcal{D}}_i}f^i(y^i;\zeta)$ are expected functions, we can not compute the full gradient $\nabla F^i (w)$.
Thus, we compute the stochastic gradient $\nabla F^i (w)$ based on some mini-batch samples. Specifically, we draw
a mini-batch samples $\mathcal{B}^i_t \subset \mathcal{D}_i$ with $|\mathcal{B}^i_t|=b$,
and then compute stochastic value of inner function $f^i_{\mathcal{B}^i_t}(w^i_{s,t})=\frac{1}{b}\sum_{j\in \mathcal{B}^i_t}f^i(w^i_{s,t};\xi_j)$ and
its stochastic gradient $\nabla f^i_{\mathcal{B}^i_t}(w^i_{s,t})=\frac{1}{b}\sum_{j\in \mathcal{B}^i_t}\nabla f^i(w^i_{s,t};\xi_j)$.
At the same time, we draw a mini-batch samples $\tilde{\mathcal{B}}^i_t \subset \tilde{\mathcal{D}}_i$ with $|\tilde{\mathcal{B}}^i_t|=b_1$,
and compute stochastic gradient of outer function $\nabla g^i_{\tilde{\mathcal{B}}^i_t}(f^i_{\mathcal{B}^i_t}(w^i_{s,t}))= \frac{1}{b_1}\sum_{j\in \tilde{\mathcal{B}}^i_t}g^i(f^i_{\mathcal{B}^i_t}(w^i_{s,t});\zeta_j)$. Finally, we obtain the stochastic gradient estimator $u^i_{s,t} = \nabla g^i_{\tilde{\mathcal{B}}^i_t}(f^i_{\mathcal{B}^i_t}(w^i_{s,t}))^T\nabla f^i_{\mathcal{B}^i_t}(w^i_{s,t})$.
Clearly, the gradient estimator $u^i_{s,t}$ is a biased estimator of $\nabla F^i(w)$, i.e., $\mathbb{E}[u^i_{s,t}] \neq \nabla F^i(w)$.

When the outer functions $\{g^i\}_{i=1}^n$ in the problem (\ref{eq:1}) are determinate and $g^1=g^2=\cdots=g^n$,
i.e., $F^i(w)=g(\mathbb{E}[f^i(w;\xi)])$, we only compute stochastic gradient $u^i_{s,t} = \nabla g(f^i_{\mathcal{B}^i_t}(w^i_{s,t}))^T\nabla f^i_{\mathcal{B}^i_t}(w^i_{s,t})$. For example, by the following Lemma \ref{lem:1}, we can obtain the simple composition problem
(\ref{eq:6}) is similarly equivalent to the minimax problem (\ref{eq:4}) with $\phi(r,1/n)= \sum_{i=1}^n r_i\log(nr_i)$.
To solve the problem (\ref{eq:6}), we can obtain
$u^i_t =\big(\exp\big(f^i_{\mathcal{B}^i_t}(w^i_t)/\gamma\big)/\gamma\big)\nabla f^i_{\mathcal{B}^i_t}(w^i_t)$.

When obtaining the stochastic gradient $u^i_{s,t}$, we will use the
SGD to update the parameter $w$ given at the line 13 in Algorithm \ref{alg:1}, where
we can choose an appropriate learning rate $\eta>0$ to guarantee the convergence of our algorithm.
In Algorithm \ref{alg:1}, we apply the local-SGD framework to reduce communication. Specifically,
our ComFedL algorithm performs multiple stochastic gradient descent (SGD) in the available devices
before communicating to the server, which can reduce the total amount of communication. In Algorithm \ref{alg:1}, we use the
synchronization gap $\tau>0$ to control the communication cost
and performances of our algorithm.

At the line 17 of Algorithm \ref{alg:1}, the server averages the local model variables $\bar{w}_{s+1} = \frac{1}{m}\sum_{i\in \mathcal{C}_s} w^i_{s,\tau}$ as in the robust FL algorithms \cite{deng2020distributionally} instead of the re-weighting formulation $\bar{w}_{s+1} = \sum_{i\in \mathcal{C}_s} r_i w^i_{s,\tau}$ in the FedAvg algorithm \cite{mcmahan2017communication}. Because our  ComFedL solves the Distributionally Robust FL as a simpler compositional optimization problem, which is equivalent to a regularized minimax problem that searches for a global variable from the worst-case loss.

\section{ Theoretical Analysis}
\label{sec:4}
In this section, we first give a key lemma that shows the robust FL can be
formed as a simple yet effective compositional optimization problem, and then provide the detailed  convergence analysis of our ComFedL algorithm.

\subsection{A Key Lemma}
In this subsection, we introduce a useful lemma that shows the robust FL can be
formed as a simple yet effective compositional optimization problem.
\begin{lemma} \label{lem:1}
The above minimax problem (\ref{eq:4}) with $\phi(r,1/n)= \sum_{i=1}^n r_i\log(nr_i)$, i.e.,
\begin{align} \label{eq:B01}
 & \min_{w \in \mathbb{R}^d} \max_{r\in \Lambda_n} \sum_{i=1}^n r_i f^i(w) - \gamma\sum_{i=1}^n r_i\log(nr_i), \\
 & \quad \mbox{s.t.} \  \Lambda_n=\big\{ r\in \mathbb{R}^n_+ \ | \ \sum_{i=1}^nr_i =1 \big\} \nonumber
\end{align}
is equivalent to the above composition problem (\ref{eq:5}), i.e.,
\begin{align} \label{eq:B02}
  \min_{w \in \mathbb{R}^d} \gamma\log\bigg( \frac{1}{n}\sum_{i=1}^n \exp\big(f^i(w)/\gamma\big)\bigg).
\end{align}
\end{lemma}

\begin{proof}
We maximize the above problem (\ref{eq:B01}) over $r\in \Lambda_n=\big\{ r\in \mathbb{R}^n_+|\sum_{i=1}^nr_i =1 \big\}$, i.e.,
\begin{align}
 \max_{r\in \Lambda_n } F(w,r) := \sum_{i=1}^nr_if^i(w)-\gamma\sum_{i=1}^nr_i\log(r_in).
\end{align}
For function $F(w,r)$, we first introduce its Lagrange function
\begin{align}
 L(w,r,\lambda)  = \sum_{i=1}^nr_if^i(w)\!-\!\gamma\sum_{i=1}^nr_i\log(r_in)  \!+\! \lambda\big( \sum_{i=1}^nr_i \!-\! 1 \big), \nonumber
\end{align}
where $\lambda$ is a Lagrange multiplier.
Then we have
\begin{align} \label{eq:B1}
   \frac{\partial L(w,r,\lambda)}{\partial r_i} = f^i(w) - \gamma\log(r_in) - \gamma + \lambda = 0.
\end{align}
According to the equality (\ref{eq:B1}), we have
\begin{align} \label{eq:B2}
 f^i(w) - \gamma\log(nr_i) = \gamma - \lambda.
\end{align}
Then by using the equality (\ref{eq:B2}), we can obtain
\begin{align}
 r_i = \frac{1}{n}\exp(\frac{\lambda}{\gamma}-1)\exp(f^i(w)/\gamma)
\end{align}
According to $\sum_{i=1}^nr_i=1$, we have
\begin{align}
 \sum_{i=1}^nr_i = \frac{1}{n}\exp(\frac{\lambda}{\gamma}-1)\sum_{i=1}^n\exp(f^i(w)/\gamma) = 1,
\end{align}
Clearly, we can obtain
\begin{align}
 \exp(1-\frac{\lambda}{\gamma}) = \frac{1}{n}\sum_{i=1}^n\exp(f^i(w)/\gamma).
\end{align}
Taking the logarithm of its both sides, we have
\begin{align} \label{eq:B3}
1-\frac{\lambda}{\gamma} = \log\big(\frac{1}{n}\sum_{i=1}^n\exp(f^i(w)/\gamma)\big).
\end{align}

According to $\sum_{i=1}^nr_i=1$ and the above equalities (\ref{eq:B2}) and (\ref{eq:B3}), we have
\begin{align}
 F(w,r) & = \sum_{i=1}^n r_i\big(f^i(w) - \gamma\log(nr_i)\big)= \gamma - \lambda \nonumber \\
 & = \gamma\log\big(\frac{1}{n}\sum_{i=1}^n\exp(f^i(w)/\gamma)\big),
\end{align}
where the last equality is due to the above equality (\ref{eq:B3}).

Thus, the above minimax problem (\ref{eq:B01}) is equivalent to the above composition problem (\ref{eq:B02}).
In the other word, by exactly maximizing the above problem (\ref{eq:4}) over $r\in \Lambda_n$,
the above minimax problem (\ref{eq:4}) with $\phi(r,1/n)= \sum_{i=1}^n r_i\log(nr_i)$ is equivalent to the composition problem (\ref{eq:5}).

\end{proof}

\subsection{ Some Mild Assumptions }
In this subsection, we introduce some mild conditions.
\begin{assumption}
There exist constants $L_f$, $L_g$ and $L$ for $\nabla f^i(w;\xi)$, $\nabla g^i(y;\zeta)$ and $\nabla F(w)$ satisfying that
for $i\in [n]$
\begin{align}
& \|\nabla f^i(w_1;\xi)\!-\!\nabla f^i(w_2;\xi)\| \!\leq \! L_f \|w_1\!-\!w_2\|, \forall w_1,w_2 \in \mathbb{R}^d,  \nonumber\\
& \|\nabla g^i(y_1,\zeta)-\nabla g^i(y_2,\zeta)\| \leq L_g \|y_1-y_2\|, \forall y_1,y_2 \in \mathbb{R}^p,   \nonumber\\
& \|\nabla F(w_1) - \nabla F(w_2)\| \leq L \|w_1-w_2\|, \nonumber
\end{align}
and the last inequality follows
\begin{align}
 F(w_2) \leq F(w_1) + \nabla F(w_1)^T(w_2-w_1) + \frac{L}{2}\|w_1-w_2\|^2.  \nonumber
\end{align}
\end{assumption}
\begin{assumption}
Gradient $\nabla g^i(y)$ and Jacobian matrix $\nabla f^i(w)$ have the upper bounds $G_g$ and $G_f$, respectively, i.e.,
for $i\in [n]$
\begin{align}
 \|\nabla f^i(w;\xi)\| \leq G_f,  \forall w \in \mathbb{R}^d;
 \|\nabla g^i(y;\zeta)\| \leq G_g,  \forall y \in \mathbb{R}^p. \nonumber
\end{align}
\end{assumption}
\begin{assumption} The variances of stochastic gradient or value of functions $f^i(w;\xi)$ and
$g^i(y;\zeta)$,  i.e., we have for all $i\in [n]$
\begin{align}
 & \mathbb{E}\|\nabla f^i(w;\xi)-\nabla f^i(w)\|^2 \leq \sigma_1^2, \nonumber \\
 & \mathbb{E}\|f^i(w;\xi)- f^i(w)\|^2 \leq \sigma_2^2, \ \forall w\in \mathbb{R}^d \nonumber \\
 & \mathbb{E}\|\nabla g^i(y;\zeta)-\nabla g^i(y)\|^2 \leq \sigma_3^2, \ \forall y\in \mathbb{R}^p \nonumber
\end{align}
where $\sigma_1,\sigma_2,\sigma_3>0$. Let $\sigma=\max(\sigma_1,\sigma_2,\sigma_3)$.
\end{assumption}
\begin{assumption}
$F(w)$ is lower bounded, i.e.,  $F^* = \inf_{w\in \mathbb{R}^d} F(w)$.
\end{assumption}

Assumptions 4.2-4.5 have been commonly used in the convergence
analysis of the composition stochastic algorithms \cite{wang2017stochastic,wang2017accelerating}. Specifically,
Assumption 4.2 ensures the smoothness of functions $ f^i(w;\xi)$, $g^i(y;\zeta)$ and $F(w)$.
Assumption 4.3 ensures the bounded gradients (or Jacobian matrix) of functions $ f^i(w;\xi)$ and $g^i(y;\zeta)$.
Assumption 4.4 ensures the bounded variances of stochastic gradient or value of functions $f^i(w;\xi)$ and
$g^i(y;\zeta)$.
Assumption 4.5 guarantees the feasibility of the problem (\ref{eq:1}).
For the special problem (\ref{eq:1}), where the outer functions $\{g^i\}_{i=1}^n$ are determinate and $g^1=\cdots=g^n$,
we only assume $\|\nabla g(y_1)-\nabla g(y_2)\| \leq L_g \|y_1-y_2\|, \ \forall y_1,y_2 \in \mathbb{R}^p$ and
$\|\nabla g(y)\| \leq G_g, \ \forall y \in \mathbb{R}^p$. For example, in the above problem (\ref{eq:6}),
$g(y), \ y\in \mathcal{Y}$ is a monotonically increasing function, where $\mathcal{Y}$ is the range of functions $\{f^i(w)\}$.
Thus, $g(y)$ is generally smooth and $\|\nabla g(y)\|$ is bounded.
\textbf{Note that} we analyze the convergence properties of our ComFedL algorithm under non-i.i.d. and nonconvex setting.
Although our convergence analysis relies on the bounded gradients in Assumption 4.4,
the exiting convergence analysis of FL algorithms \cite{li2019convergence,zhang2020fedpd,chen2020fedcluster,deng2020distributionally}
under non-i.i.d. and nonconvex (or strongly convex)
setting also rely on
the bounded gradients.

\subsection{ Convergence Analysis }
In this subsection, we detail the convergence analysis of
our CompFedL algorithm. In Algorithm \ref{alg:1}, we only obtain a \textbf{biased} stochastic gradient in each device,
due to the compositional loss function in our new FL framework.
While the \textbf{unbiased} stochastic gradient is easily obtained in each device in the existing FL framework.
Thus, our convergence analysis can not easily follow the existing convergence analysis of the FL \cite{li2019convergence,khaled2020tighter,zhang2020fedpd}.
For notational simplicity, let
\begin{align}
 \bar{w}_{s,t} & = \frac{1}{m}\sum_{i\in \mathcal{C}_s} w^i_{s,t}, \nonumber \\
 \bar{u}_{s,t} & = \frac{1}{m}\sum_{i\in \mathcal{C}_s} u^i_{s,t}
  = \frac{1}{m}\sum_{i\in \mathcal{C}_s}\nabla g^i_{\tilde{\mathcal{B}}^i_t}(f^i_{\mathcal{B}^i_t}(w^i_{s,t}))^T\nabla f^i_{\mathcal{B}^i_t}(w^i_{s,t}), \nonumber
\end{align}
then we have
$\bar{w}_{s,t+1} = \bar{w}_{s,t} - \eta\bar{u}_{s,t}
$. Let
\begin{align} \label{eq:A1}
 U_{s,0} = \frac{1}{m}\sum_{i\in \mathcal{C}_s} \nabla F^i(\bar{w}_{s}) = \frac{1}{m}\sum_{i\in \mathcal{C}_s} \nabla g^i\big(f^i(\bar{w}_{s})\big)^T\nabla f^i(\bar{w}_{s}),
\end{align}
we have $\mathbb{E}[U_{s,0}] = \nabla F(\bar{w}_s)$.

\begin{lemma} \label{lem:A1}
Under the about Assumptions, we have
\begin{align}
\mathbb{E}\|\bar{w}_{s,t} - \bar{w}_s\|^2 \leq \tau^2 \eta^2 G^2_gG^2_f. \nonumber
\end{align}
\end{lemma}
\begin{proof}
Since $\bar{w}_s = w^i_{s,0}$, we have
\begin{align}
 & \mathbb{E}\|\bar{w}_{s,t} - \bar{w}_s\|^2 = \mathbb{E}\|\frac{1}{m}\sum_{i\in \mathcal{C}_s}w^i_{s,t} - \bar{w}_s\|^2  \nonumber \\
 & \leq \frac{1}{m}\sum_{i\in \mathcal{C}_s}\mathbb{E}\|w^i_{s,t} - \bar{w}_s\|^2 = \frac{1}{m}\sum_{i\in \mathcal{C}_s}\mathbb{E} \|w^i_{s,t} - w^i_{s,0}\|^2 \nonumber \\
 & = \frac{1}{m}\sum_{i\in \mathcal{C}_s}\mathbb{E} \|\sum_{j=0}^{t-1}\eta u^i_{s,j}\|^2 \nonumber \\
 & \leq \frac{1}{m}\sum_{i\in \mathcal{C}_s} \bigg(t \eta^2 \sum_{j=0}^{t-1}\mathbb{E}\|\nabla g_{\tilde{\mathcal{B}}^i_t}(f^i_{\mathcal{B}^i_t}(w^i_{s,t}))^T\nabla f^i_{\mathcal{B}^i_t}(w^i_{s,t})\|^2 \bigg) \nonumber \\
 & \leq t^2 \eta^2 G^2_gG^2_f \leq \tau^2 \eta^2 G^2_gG^2_f,
\end{align}
where the second inequality holds by Assumption 2.
\end{proof}

\begin{lemma} \label{lem:A2}
Under the above Assumptions, we have
\begin{align}
&\mathbb{E}\|\bar{u}_{s,t}-U_{s,0}\|^2 \leq 5(G^2_gL^2_f+G^4_fL^2_g)\tau^2 \eta^2 G^2_gG^2_f \nonumber \\
& \quad + \frac{5G^2_f\sigma^2}{b_1}
 + \frac{5G^2_g\sigma^2}{b}  + \frac{5L_g^2G_f^2\sigma^2}{b}.
\end{align}
\end{lemma}
\begin{proof}
By using the above equality \eqref{eq:A1}, we have
\begin{align}
&\mathbb{E}\|\bar{u}_{s,t}-U_{s,0}\|^2  \nonumber \\
& =   \mathbb{E}\| \frac{1}{m} \sum_{i\in \mathcal{C}_s}  \nabla g^i_{\tilde{\mathcal{B}}^i_t}(f^i_{\mathcal{B}^i_t}(w^i_{s,t}))^T\nabla f^i_{\mathcal{B}^i_t}(w^i_{s,t})  \nonumber \\
& \ -  \nabla g^i(f^i(\bar{w}_{s}))^T\nabla f^i(\bar{w}_{s}) \|^2 \nonumber \\
& \leq  \frac{1}{m} \sum_{i\in \mathcal{C}_s} \mathbb{E}\| \nabla g^i_{\tilde{\mathcal{B}}^i_t}(f^i_{\mathcal{B}^i_t}(w^i_{s,t}))^T\nabla f^i_{\mathcal{B}^i_t}(w^i_{s,t}) \nonumber \\
& \ -  \nabla g^i(f^i(\bar{w}_{s}))^T\nabla f^i(\bar{w}_{s})\|^2 \nonumber \\
& =  \frac{1}{m} \sum_{i\in \mathcal{C}_s} \mathbb{E}\| \nabla g^i_{\tilde{\mathcal{B}}^i_t}(f^i_{\mathcal{B}^i_t}(w^i_{s,t}))^T\nabla f^i_{\mathcal{B}^i_t}(w^i_{s,t})  \nonumber \\
& \ - \nabla g^i_{\tilde{\mathcal{B}}^i_t}(f^i_{\mathcal{B}^i_t}(w^i_{s,t})) ^T\nabla f^i_{\mathcal{B}^i_t}(\bar{w}_{s}) \nonumber \\
& \ + \nabla g^i_{\tilde{\mathcal{B}}^i_t}(f^i_{\mathcal{B}^i_t}(w^i_{s,t})) ^T\nabla f^i_{\mathcal{B}^i_t}(\bar{w}_{s})  \!-\! \nabla g^i(f^i_{\mathcal{B}^i_t}(w^i_{s,t})) ^T\nabla f^i_{\mathcal{B}^i_t}(\bar{w}_{s}) \nonumber \\
& \ + \nabla g^i(f^i_{\mathcal{B}^i_t}(w^i_{s,t})) ^T\nabla f^i_{\mathcal{B}^i_t}(\bar{w}_{s})- \nabla g^i(f^i_{\mathcal{B}^i_t}(w^i_{s,t})) ^T\nabla f^i(\bar{w}_{s}) \nonumber \\
& \ + \nabla g^i(f^i_{\mathcal{B}^i_t}(w^i_{s,t})) ^T\nabla f^i(\bar{w}_{s}) -\nabla g^i(f^i_{\mathcal{B}^i_t}(\bar{w}_s)) ^T\nabla f^i(\bar{w}_{s}) \nonumber \\
& \ + \nabla g^i(f^i_{\mathcal{B}^i_t}(\bar{w}_s)) ^T\nabla f^i(\bar{w}_{s}) - \nabla g^i(f^i(\bar{w}_{s})) ^T\nabla f^i(\bar{w}_{s})\|^2 \nonumber \\
& \leq \frac{5G^2_gL^2_f}{m}\sum_{i\in \mathcal{C}_s}\mathbb{E}\|w^i_{s,t} - \bar{w}_{s}\|^2 + \frac{5G^2_f\sigma^2}{b_1} + \frac{5G^2_g\sigma^2}{b} \nonumber \\
& \ + \frac{5G^4_fL^2_g}{m}\sum_{i\in \mathcal{C}_s}\mathbb{E}\|w^i_{s,t} - \bar{w}_{s}\|^2 + \frac{5L_g^2G_f^2\sigma^2}{b} \nonumber \\
& \leq 5(G^2_gL^2_f+G^4_fL^2_g)\tau^2 \eta^2 G^2_gG^2_f + \frac{5G^2_f\sigma^2}{b_1}  \nonumber \\
& \ + \frac{5G^2_g\sigma^2}{b} + \frac{5L_g^2G_f^2\sigma^2}{b},
\end{align}
where the second inequality is due to Assumptions 1-3, and the last inequality holds by Lemma
\ref{lem:A1}.
\end{proof}

\begin{theorem}
Under the above assumptions, in Algorithm 1, given $\eta>0$ and $b=b_1>0$, we have
\begin{align}
& \frac{1}{S}\sum_{s=0}^{S-1}\mathbb{E}\|\nabla F(\bar{w}_{s})\|^2 \nonumber \\
& \leq \frac{F(\bar{w}_0)-F^*}{T\eta} + G_fG_g \bigg(\sqrt{5}H\tau G_gG_f\eta \frac{\sqrt{5}G_f\sigma}{\sqrt{b_1}} \nonumber \\
& \ + \frac{\sqrt{5}G_g\sigma}{\sqrt{b}} + \frac{\sqrt{5}L_gG_f\sigma}{\sqrt{b}}\bigg) + L\tau\eta G^2_gG^2_f + \frac{L\eta}{2}G_f^2G_g^2,
\end{align}
where $T=\tau S$ and $H=\sqrt{G^2_gL^2_f+G^4_fL^2_g}$.
\end{theorem}

\begin{proof}
By the smoothness of the function $F(x)$, we have
\begin{align} \label{eq:A2}
& \mathbb{E}[F(\bar{w}_{s,t+1})] \nonumber \\
& \leq \mathbb{E}[F(\bar{w}_{s,t})]  +  \mathbb{E}\langle\nabla F(\bar{w}_{s,t}), \bar{w}_{s,t+1}-\bar{w}_{s,t}\rangle  \nonumber \\
& \quad +  \frac{L}{2}\mathbb{E}\|\bar{w}_{s,t+1}-\bar{w}_{s,t}\|^2 \nonumber \\
& = \mathbb{E}[F(\bar{w}_{s,t})] + \mathbb{E}\langle\nabla F(\bar{w}_{s,t}), -\eta \bar{u}_{s,t}\rangle
+ \frac{L}{2}\mathbb{E}\|\eta \bar{u}_{s,t}\|^2 \nonumber \\
& = \mathbb{E}[F(\bar{w}_{s,t})] - \eta\mathbb{E}\langle\nabla F(\bar{w}_{s,t}), \bar{u}_{s,t}-U_{s,0}\rangle  \nonumber \\
& \quad - \eta\mathbb{E}\langle\nabla F(\bar{w}_{s,t})-\nabla F(\bar{w}_{s}), U_{s,0}\rangle - \eta\mathbb{E}\langle\nabla F(\bar{w}_{s}), U_{s,0}\rangle \nonumber \\
& \quad + \frac{L}{2}\mathbb{E}\|\eta \bar{u}_{s,t}\|^2 \nonumber \\
& \leq \mathbb{E}[F(\bar{w}_{s,t})] + \eta\mathbb{E}\big(\|\nabla F(\bar{w}_{s,t})\|\|\bar{u}_{s,t}-U_{s,0}\|\big)  \nonumber \\
& \quad + \eta\mathbb{E}\big(\|\nabla F(\bar{w}_{s,t})-\nabla F(\bar{w}_{s})\|\| U_{s,0}\|\big) - \eta\mathbb{E}\|\nabla F(\bar{w}_{s})\|^2 \nonumber \\
& \quad + \frac{L\eta^2}{2}\mathbb{E}\|\bar{u}_{s,t}\|^2 \nonumber \\
& \leq \mathbb{E}[F(\bar{w}_{s,t})] + \eta G_fG_g\mathbb{E}\|\bar{u}_{s,t}-U_{s,0}\| + \eta LG_fG_g\mathbb{E}\|\bar{w}_{s,t}-\bar{w}_{s}\| \nonumber \\
& \quad  - \eta\mathbb{E}\|\nabla F(\bar{w}_{s})\|^2 + \frac{L\eta^2}{2}G_f^2G_g^2 \nonumber \\
& \leq \mathbb{E}[F(\bar{w}_{s,t})] + \eta G_fG_g\sqrt{\mathbb{E}\|\bar{u}_{s,t}-U_{s,0}\|^2}     \nonumber \\
& \quad + \eta LG_fG_g\sqrt{\mathbb{E}\|\bar{w}_{s,t}-\bar{w}_{s}\|^2} - \eta\mathbb{E}\|\nabla F(\bar{w}_{s})\|^2 + \frac{L\eta^2}{2}G_f^2G_g^2 \nonumber \\
& \leq \mathbb{E}[F(\bar{w}_{s,t})] + \eta G_fG_g \bigg( \sqrt{5(G^2_gL^2_f+G^4_fL^2_g)}\tau\eta G_gG_f
\nonumber \\
& \quad + \frac{\sqrt{5}G_f\sigma}{\sqrt{b_1}} + \frac{\sqrt{5}G_g\sigma}{\sqrt{b}} + \frac{\sqrt{5}L_gG_f\sigma}{\sqrt{b}}\bigg)   +  L\tau\eta^2 G^2_gG^2_f  \nonumber \\
& \quad - \eta\mathbb{E}\|\nabla F(\bar{w}_{s})\|^2 + \frac{L\eta^2}{2}G_f^2G_g^2,
 \end{align}
where the second inequality is due to Cauchy-Schwarz inequality and $\mathbb{E}[U_{s,0}] = \nabla F(\bar{w}_s)$; the second last inequality holds by the concavity of $\sqrt{x}$, and the last inequality is due to Lemmas \ref{lem:A1} and
\ref{lem:A2}.
Thus, we have
\begin{align}
& \eta\mathbb{E}\|\nabla F(\bar{w}_{s})\|^2 \leq \mathbb{E}[F(\bar{w}_{s,t})] - \mathbb{E}[F(\bar{w}_{s,t+1})] \nonumber \\
& \ + \eta G_fG_g \bigg( \sqrt{5(G^2_gL^2_f+G^4_fL^2_g)}\tau\eta G_gG_f \!+\! \frac{\sqrt{5}G_f\sigma}{\sqrt{b_1}} \!+\! \frac{\sqrt{5}G_g\sigma}{\sqrt{b}} \nonumber \\
& \ + \frac{\sqrt{5}L_gG_f\sigma}{\sqrt{b}}\bigg) + L\tau\eta^2 G^2_gG^2_f + \frac{L\eta^2}{2}G_f^2G_g^2.
\end{align}

Since $w_{s,0}=\bar{w}_{s}$,
telescoping the above inequality over $s=0,1,\cdots,S-1$ and $t=0,1,\cdots,\tau-1$, we have
\begin{align}
& \frac{1}{S}\sum_{s=0}^{S-1}\mathbb{E}\|\nabla F(\bar{w}_{s})\|^2  \\
& \leq \frac{F(\bar{w}_0)-F(\bar{w}_{T})}{T\eta} + G_fG_g \bigg( \sqrt{5(G^2_gL^2_f+G^4_fL^2_g)}\tau\eta G_gG_f \nonumber \\
& \ + \frac{\sqrt{5}G_f\sigma}{\sqrt{b_1}} \!+\! \frac{\sqrt{5}G_g\sigma}{\sqrt{b}} \!+\! \frac{\sqrt{5}L_gG_f\sigma}{\sqrt{b}}\bigg) \!+\! L\eta G^2_gG^2_f\tau \!+\! \frac{L\eta}{2}G_f^2G_g^2 \nonumber \\
& \leq \frac{F(\bar{w}_0)-F^*}{T\eta} + G_fG_g \bigg( \sqrt{5(G^2_gL^2_f+G^4_fL^2_g)}\tau\eta G_gG_f     \nonumber \\
& \ + \frac{\sqrt{5}G_f\sigma}{\sqrt{b_1}} \!+\! \frac{\sqrt{5}G_g\sigma}{\sqrt{b}} \!+\! \frac{\sqrt{5}L_gG_f\sigma}{\sqrt{b}} \bigg) \!+\! L\tau\eta G^2_gG^2_f \!+\! \frac{L\eta}{2}G_f^2G_g^2, \nonumber
\end{align}
where the last inequality is due to Assumption 4.
\end{proof}

\begin{remark}
Let $\eta=\frac{1}{T^{\alpha_1}}, \ 0<\alpha_1\leq 1$, $b = \frac{1}{T^{\alpha_2}}, \ \alpha_2>0$ and $b_1 = \frac{1}{T^{\alpha_3}}, \ \alpha_3>0$, we have
$\frac{1}{S}\sum_{s=0}^{S-1}\mathbb{E}\|\nabla F(\bar{w}_{s})\|^2 \leq O(\frac{1}{T^{1-\alpha_1}} + \frac{\tau}{T^{\alpha_1}} + \frac{1}{T^{0.5\alpha_2}} + \frac{1}{T^{0.5\alpha_3}} )$.
Thus, our ComFedL algorithm has a convergence rate of $O(\frac{1}{T^{1-\alpha_1}}+ \frac{\tau}{T^{\alpha_1}} + \frac{1}{T^{0.5\alpha_2}} + \frac{1}{T^{0.5\alpha_3}})$.
When $\alpha_1 = \frac{1}{2}$, $\alpha_2 = \alpha_3=1$ and $\tau=O(1)$, our ComFedL algorithm reaches a convergence rate of $O(\frac{1}{\sqrt{T}})$.
\end{remark}

\begin{figure}[h]
  \centering
  \includegraphics[width=0.48\linewidth]{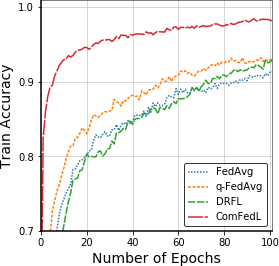}
  \includegraphics[width=0.48\linewidth]{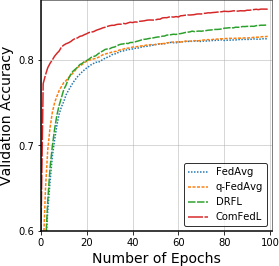}
  \includegraphics[width=0.48\linewidth]{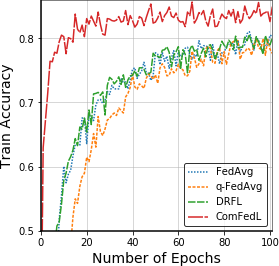}
  \includegraphics[width=0.48\linewidth]{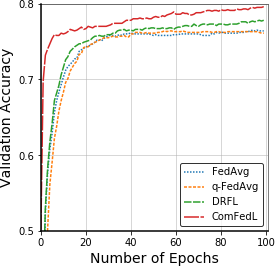}
  \caption{Train and Validation accuracy over imbalanced MNIST dataset with different robust FL methods. The top figures show average accuracy, and the bottom figures show the worst accuracy.}
\label{fig:comp1}
\vspace{-0.5cm}
\end{figure}

\begin{figure}[h]
  \centering
  \includegraphics[width=0.48\linewidth]{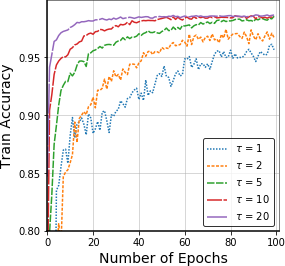}
  \includegraphics[width=0.48\linewidth]{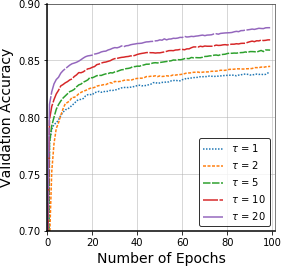}
  \includegraphics[width=0.48\linewidth]{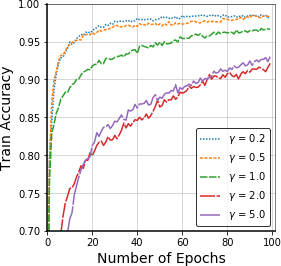}
  \includegraphics[width=0.48\linewidth]{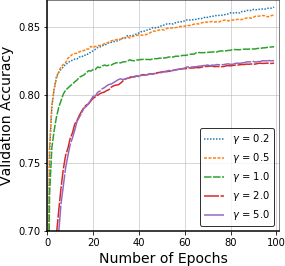}
  \caption{Comparing the effects of synchronization gap $\tau$ (top row) and regularization parameter $\gamma$ (bottom row) to our algorithm.}
\label{fig:local-iter-gamma-comp1}
\vspace{-0.5cm}
\end{figure}

\section{Experiments}
In this section, we empirically validate the efficacy of our ComFedL algorithm. Specifically, we apply our algorithm to two tasks
\emph{i.e.} Distributionally Robust Federated Learning
and Distribution Agnostic MAML-based Personalized Federated Learning.
In the experiments, all algorithms are implemented based on the \textit{Distributed} Library of Pytorch~\cite{paszke2019pytorch},
and all experiments are run over a machine with 4 NVIDIA P40 GPUs.

\subsection{Distributionally Robust Federated Learning}
In this subsection, we conduct the distributionally robust FL task to verify
effectiveness of our algorithm.
In this set of experiments, we consider the multi-class classification problem over the MNIST~\cite{lecun2010mnist} dataset with logistic regression model.
The experiments are run over 10 clients and 1 server. In particular, we randomly select one client to have 5000 images,
while the remained clients are distributed much less data, which is 20 images. This way of data construction is aimed to create imbalance among different client's datasets.
To get a good performance over this type of imbalanced datasets, the algorithm needs to attach more importance over the hardest task, \emph{i.e.} the client that has dominant number of images.
So it is a good way to test the ability of our algorithm to adapt to such imbalance.

We compare our algorithm with several state-of-the-art distributionally robust federated learning algorithms.
In particular, we compare with DRFL~\cite{deng2020distributionally}, q-FedAvg~\cite{li2019fair} and the basic FedAvg~\cite{mcmahan2017communication}.
In experiments, we set synchronization gap as 5 for all methods.
As for other hyper-parameters, we perform grid search to find the optimal ones for each method and the search space is included in the following subsection \ref{sec:5.3}.
More specifically, Learning rate is 0.01 for all methods. In our algorithm, regularization parameter $\gamma$ is 0.2.
In DRFL, client weights learning rate $\gamma$ is $8\times10^{-2}$. In q-FedAvg, $q$ is set as 0.2. Finally, the synchronization gap is 5 by default.
For all ablation studies, the hyper-parameters are chosen as above if not specified.

We show the result in Figure~\ref{fig:comp1}, where we report both average accuracy and the worst accuracy (among all the clients).
As shown by the figure, our algorithm significantly outperforms all baselines in both metrics.
Note that the FedAvg algorithm assigns weights proportional to the number of samples at each client,
which might lead to bad validation performance for clients with smaller number of images (\emph{e.g.} clients with only 20 images).
Furthermore, DRFL solves a challenging minimax problem by optimizing the weight $r_i$ explicitly, while our algorithm can dynamically adjust the weight of clients,
which makes our model simpler to train.
Then in Figure~\ref{fig:local-iter-gamma-comp1}, we show the robustness of our algorithm by varying the synchronization gap $\tau$
and the regularization parameter $\gamma$. As shown by the figure, our algorithm can get good train and validation accuracy with different values of $\tau$,
furthermore, the algorithm converges much faster when we increase $\tau$,
which shows the effectiveness of running multiple local epochs. Next, we show the effects of $\gamma$. Recall that $\gamma$ is a
regularization parameter that penalizes the divergence $\phi(r,1/n)$. The larger $\gamma$ is,
the more the algorithm emphasizes on getting $r_i$ close to the the average weight $\frac{1}{n}$. Since in our data-set,
the optimal weights are far from $\frac{1}{n}$, we should pick $\gamma$ relatively small. This is verified by the results in Figure~\ref{fig:local-iter-gamma-comp1}.
The algorithm converges much faster when we choose $\gamma = 0.1/0.2$ compared to that of $2/5$ in terms of both train and validation accuracy.

\begin{figure}[t]
  \centering
  \includegraphics[width=0.48\linewidth]{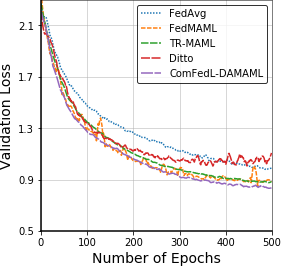}
  \includegraphics[width=0.48\linewidth]{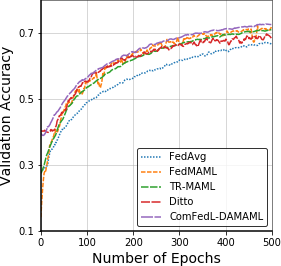}
  \caption{Validation loss (left) and Validation accuracy (right) over heterogeneous CIFAR-10 dataset with different personalized FL methods.}
\label{fig:comp_fedmaml}
\vspace{-0.7cm}
\end{figure}

\subsection{Distribution Agnostic MAML-based Personalized Federated Learning }
In this subsection, we conduct Distribution Agnostic MAML-based personalized federated learning task to verify efficacy of our algorithm. Here we let ComFedL-DAMAML denote our algorithm for solving the above problem (\ref{eq:10}), i.e., a distributionally robust MAML-based personalized FL problem with regularization problem. We compare with the following baselines: FedAvg~\cite{mcmahan2017communication}, FedMAML~\cite{fallah2020personalized}, TR-MAML~\cite{collins2020distribution} and Ditto~\cite{li2021ditto}.
In this set of experiments, we consider the multi-class classification task over the CIFAR10~\cite{krizhevsky2009learning} dataset with a 4-layer Convolutional Neural Network (CNN) used  in~\cite{finn2017model}.
We create the heterogeneous training (validation) dataset as follows: we create 10 clients and 1 server, while the dataset
over each client includes images from a dominant class\footnote{A dominant class is the class with most samples over a client.
In our experiments, each client has a different dominant class, e.g. client 1 has 60\% samples from the airplane class and the remained 40\% samples include other classes.
The percentage of number of samples from a dominant class over each client is denoted by the hyper-parameter $\rho$.} and a small percentage of images from other classes.
The dominant class is different for different clients. More precisely, client-$i$ owns $\rho$ percentage images of class-$i$,
and $(1 -\rho)/9$ for other classes. For $\rho > 0.1$ the images of each client will be dominated by a different class. The data distribution of each client is heterogeneous by construction, so it brings extra benefit by tuning a personalized model over each client.
For all methods, we perform grid search to find the optimal hyper-parameters and the search space is reported in the following subsection \ref{sec:5.3}.
For FedAvg, the learning rate is 0.1; For FedMAML, both the inner and outer learning rate is 0.1;
For TR-MAML, the inner learning rate is 0.05 and the outer learning rate is 0.1, while the learning rate of client weights is 0.08;
For Ditto, the optimal learning rate for both the global and local models are 0.2, and the regularization parameter $\lambda$ is 0.1.
For our ComFedL-DAMAML, the inner learning rate is set as 0.05, outer learning rate is 0.1 and the regularization parameter $\gamma$ is 0.5.
Finally, the heterogeneity parameter $\rho$ is 0.28 and the synchronization gap is 5 by default. For all ablation studies, the hyper-parameters are chosen as above if not specified.

\begin{figure}[t]
  \centering
  \includegraphics[width=0.48\linewidth]{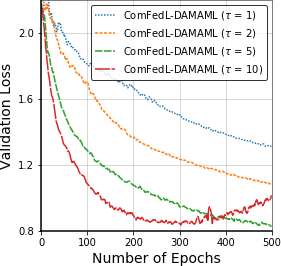}
  \includegraphics[width=0.48\linewidth]{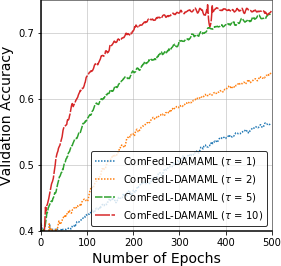}
  \includegraphics[width=0.48\linewidth]{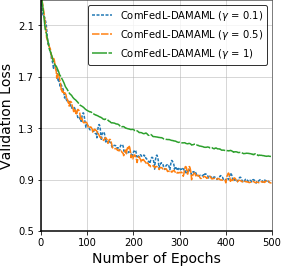}
  \includegraphics[width=0.48\linewidth]{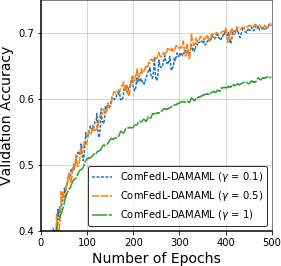}
  \caption{Comparing the effects of synchronization gap $\tau$ (top row) and regularization parameter $\gamma$ (bottom row) over ComFedL-DAMAML.}
\label{fig:comp_fedmaml_local}
\vspace{-0.5cm}
\end{figure}

\begin{figure}[h]
  \centering
  \includegraphics[width=0.48\linewidth]{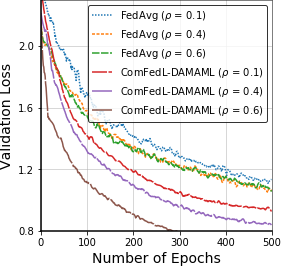}
  \includegraphics[width=0.48\linewidth]{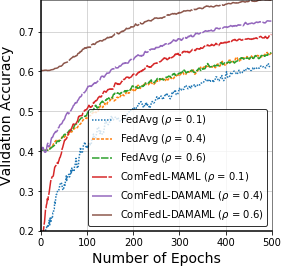}
  \caption{Comparing the effects of data heterogeneity $\rho$ over ComFedL-DAMAML.}
\label{fig:comp_fedmaml_rho}
\vspace{-0.4cm}
\end{figure}



In Figure~\ref{fig:comp_fedmaml}, we compare ComFedL-DAMAML with other baseline methods. For fair comparison, FedAvg is trained with one step at test time. As shown by the figures, ComFedL-DAMAML outperforms other baselines. The basic FedAvg algorithm can not adapt to the heterogeneity of clients well and get the worst validation loss (accuracy). FedMAML and Ditto perform better but the loss (accuracy) curve is very noisy. TR-MAML gets a smoother curve compared to FedMAML due to its better balance of different clients. While our method outperforms all these methods. Our ComFedL-DAMAML algorithm adaptively adjusts the weight of clients based on the task's performance (training loss). In other words, if the data distribution of a client is hard to learn (higher training loss), the algorithm increases its learning rate, while for clients with easier distributions, the learning rate is decreased. In summary, the results show that our ComFedL-DAMAML can also accelerate the personalized FL.

What's more, we also test the effect of synchronization gap $\tau$ as shown in the top row of Figure~\ref{fig:comp_fedmaml_local}. As shown in the figure, ComFedL-DAMAML converges much faster when we increase $\tau$, this shows the effectiveness of running multiple local epochs. Then in the bottom row of Figure~\ref{fig:comp_fedmaml_local}, we compare effects of different regularization parameter $\gamma$. As shown by experiments, our algorithm is pretty robust with different values of $\gamma$. Finally, we test the effect of heterogeneity coefficient $\rho$ in Figure~\ref{fig:comp_fedmaml_rho}. $\rho$ represents the data heterogeneity among the clients, the larger $\rho$ is, the greater data heterogeneity. As shown by the figure, ComFedL-MAML consistently outperforms the FedAvg method over different $\rho$, especially when $\rho$ is large.

\subsection{Hyper-parameter selection in the experiments}
\label{sec:5.3}
In the above Distributionally Robust Federated Learning experiments: For the learning rate, we search from [0.001, 0.01, 0.05, 0.1, 0.2, 0.5, 1].
We observe that when learning rate is 1, most methods just diverge. For our method, we search the regularization parameter from [0.1, 0.5, 1, 5];
For DRFL, we search the client weights learning rate from [8e-3, 8e-2, 8e-1]. For q-FedAvg, we search the $q$ value from [0.1, 0.2, 0.5, 1, 2].

In the above Distribution Agnostic MAML-based Personalized Federated Learning experiments: For the learning rate (both inner and outer if two types of learning rates are needed),
we search from [0.001, 0.01, 0.05, 0.1, 0.2, 0.5, 1]. For our method, we search the regularization parameter from [0.1, 0.5, 1, 5];
For TR-MAML, we search the client weights learning rate from [8e-3, 8e-2, 8e-1]; For Ditto, we search the regularization parameter $\lambda$ from [0.05, 0.1, 0.5, 1, 5]

\section{Conclusion}
In the paper, we introduced a new compositional FL framework, and proposed an effective and efficient compositional FL (ComFedL) algorithm for solving this
compositional FL framework.
To the best of our knowledge, our new compositional FL is the
first work to bridge federated learning with composition stochastic optimization. In particular, we first transform the distributionally robust federated learning
(i.e., a minimax problem) into a simple composition problem by using KL divergence regularization.


%

%

%

\ifCLASSOPTIONcaptionsoff
  \newpage
\fi



%
%
%

\bibliographystyle{IEEEtran}
\bibliography{IEEEabrv,CompFL}

\end{document}